\documentclass[conference]{IEEEtran}
\IEEEoverridecommandlockouts
\usepackage{cite}
\usepackage{amsmath,amssymb,amsfonts}
\usepackage{algorithmic}
\usepackage{graphicx}
\usepackage{textcomp}
\usepackage{xcolor}
\usepackage{tcolorbox}
\usepackage{tikz}
\usepackage{pgfplots, pgfplotstable}
\usepgfplotslibrary{colormaps, groupplots}
\pgfplotsset{compat=1.18}
\usepackage{array}

\def\BibTeX{{\rm B\kern-.05em{\sc i\kern-.025em b}\kern-.08em
    T\kern-.1667em\lower.7ex\hbox{E}\kern-.125emX}}

\makeatletter
 \let\old@ps@headings\ps@headings
 \let\old@ps@IEEEtitlepagestyle\ps@IEEEtitlepagestyle
 \def\confheader#1{%
 \def\ps@IEEEtitlepagestyle{%
 \old@ps@IEEEtitlepagestyle%
 \def\@oddhead{\strut\hfill#1\hfill\strut}%
 \def\@evenhead{\strut\hfill#1\hfill\strut}%
 }%
 \ps@headings%
 }
 \makeatother

\confheader{%
 2024 Second International Conference on Data Science and Information System (ICDSIS)
 }

 \usepackage[pscoord]{eso-pic}
\newcommand{\placetextbox}[3]{
 \setbox0=\hbox{#3}
 \AddToShipoutPictureFG*{ \put(\LenToUnit{#1\paperwidth},\LenToUnit{#2\paperheight}){\vtop{{\null}\makebox[0pt][c]{#3}}}
 }
 }
 \placetextbox{.23}{0.055}{\small{979-8-3503-8166-5/24/\$31.00~\copyright 2024 IEEE}}

\begin{document}

\title{Evaluating Consistency and Reasoning Capabilities of Large Language Models} 

\author{\IEEEauthorblockN{1\textsuperscript{st} Yash Saxena}
\IEEEauthorblockA{\textit{School of Computing Science}  \\
\textit{and Engineering}\\
\textit{Galgotias University}\\
Greater Noida, India \\
yashsaxena2111@gmail.com}
\and
\IEEEauthorblockN{2\textsuperscript{nd} Sarthak Chopra}
\IEEEauthorblockA{\textit{School of Computing Science} \\
\textit{and Engineering}\\
\textit{Galgotias University}\\
Greater Noida, India\\
sc1692002@gmail.com}
\and
\IEEEauthorblockN{3\textsuperscript{rd} Arunendra Mani Tripathi}
\IEEEauthorblockA{\textit{School of Computing Science} \\
\textit{and Engineering}\\
\textit{Galgotias University}\\
Greater Noida, India \\
arunendra.tripathi@galgotiasuniversity.edu.in}}


\maketitle

\begin{abstract}
Large Language Models (LLMs) are extensively used today across various sectors, including academia, research, business, and finance, for tasks such as text generation, summarization, and translation. Despite their widespread adoption, these models often produce incorrect and misleading information, exhibiting a tendency to hallucinate. This behavior can be attributed to several factors, with consistency and reasoning capabilities being significant contributors. LLMs frequently lack the ability to generate explanations and engage in coherent reasoning, leading to inaccurate responses. Moreover, they exhibit inconsistencies in their outputs. This paper aims to evaluate and compare the consistency and reasoning capabilities of both public and proprietary LLMs. The experiments utilize the Boolq dataset as the ground truth, comprising questions, answers, and corresponding explanations. Queries from the dataset are presented as prompts to the LLMs, and the generated responses are evaluated against the ground truth answers. Additionally, explanations are generated to assess the models' reasoning abilities. Consistency is evaluated by repeatedly presenting the same query to the models and observing for variations in their responses. For measuring reasoning capabilities, the generated explanations are compared to the ground truth explanations using metrics such as BERT, BLEU, and F-1 scores. The findings reveal that proprietary models generally outperform public models in terms of both consistency and reasoning capabilities. However, even when presented with basic general knowledge questions, none of the models achieved a score of 90\% in both consistency and reasoning. This study underscores the direct correlation between consistency and reasoning abilities in LLMs and highlights the inherent reasoning challenges present in current language models.      
\end{abstract}

\begin{IEEEkeywords}
LLM, Reasoning, Consistency, Explanability, Prompting
\end{IEEEkeywords}

\section{Introduction}

In recent years, Large Language Models (LLMs) have gained a lot of traction across various demographics including different age groups, genders, and cultures. The advancements in Generative Artificial Intelligence (also known as GenAI) have made LLMs easily accessible to the general public. These models find applications in text generation, translation, summarization, and more, with people from diverse sectors such as academia, finance, research, legislature, and business using them extensively.

LLMs have become valuable tools for tasks like obtaining detailed information on specific topics, composing emails, and overall reducing the workload of people. However, it is crucial to exercise caution when relying on these models. Despite their capabilities, LLMs are not infallible and may occasionally produce inaccurate or hallucinated information. Blindly trusting these models without verification can lead to potentially dangerous outcomes. Therefore, it is essential for users to approach LLM-generated content with a critical mindset and verify information when necessary.

The hallucination tendency of these models can be attributed to factors like inadequate information on a given topic, lack of reasoning capabilities, lack of consistency in responses etc. These models often struggle to offer strong, explainable reasons for their responses to queries, highlighting a deficiency in their reasoning capabilities \cite{huang2022towards}.

\begin{figure*}[htbp]
\centerline{\includegraphics[width=1.0\linewidth]{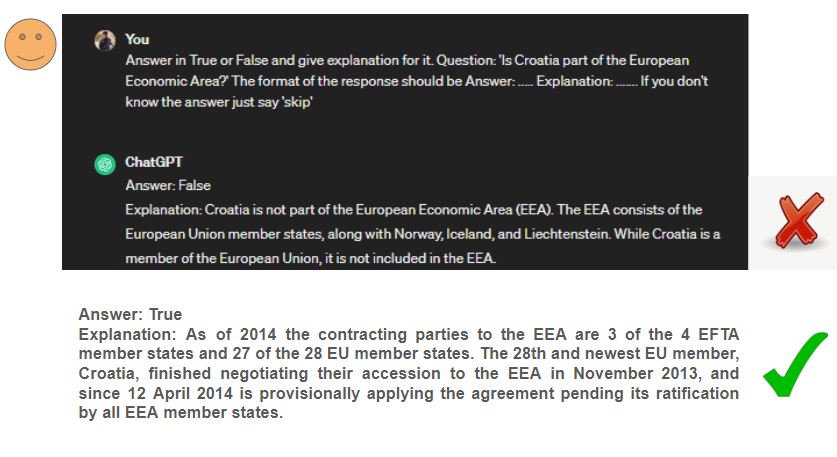}}
\caption{Example showcasing how ChatGPT (GPT-3.5) gives incorrect answer with incorrect reasoning on a query taken from the Boolq Dataset. The correct answer with explanation taken from the dataset is also provided}
\label{fig}
\end{figure*}

From Fig. \ref{fig}, it is evident that state-of-the-art models, such as ChatGPT (GPT-3.5), not only provide incorrect answers but also accompany them with fabricated explanations that support these inaccuracies. This observation underscores the continued challenge that large language models face in reasoning effectively, as they tend to generate a significant amount of hallucinated information even when confronted with general knowledge-based questions. The correct answer, accompanied by an appropriate explanation, is also presented in the figure for reference.

In this study, the reasoning capabilities and consistency of various proprietary and public LLMs on the Boolq dataset \cite{b2} are evaluated. The Boolq dataset comprises general true and false type questions along with explanations for each answer. The evaluation involves comparing the explanations provided in the dataset with those generated by the LLMs to gauge their reasoning capabilities. Additionally, each answer is generated three times to analyze the consistency of the LLMs' responses.

For the experiments, the use of both public and proprietary LLMs is made. In case of proprietary LLMs the use OpenAI's GPT-4-Turbo, GPT-4, and GPT-3.5-Turbo is made. And in case of public LLMs the use LLAMA-2-7b \cite{b3}, Mistral-7b \cite{b4} and Mixtral-8x7b \cite{b5} is made.  

\section{Related Works}

\subsection{Hallucinations in LLMs}

The issue of LLMs providing inaccurate information or hallucinating is widely acknowledged in the literature \cite{b6}. In \cite{b7}, the inevitability of hallucination in LLMs is substantiated. Furthermore, \cite{b8} demonstrates that hallucinations in LLMs can be predominantly attributed to inaccuracies in training data. The categorization of hallucinations in LLMs into different types is addressed in \cite{b9}. In \cite{b10}, the authors propose a technique for detecting hallucinations in LLMs. Moving a step further, \cite{b11} focuses on detecting hallucination in Retrieval Augmented Generation (RAG) trained LLMs. Additionally, \cite{b12} introduces a benchmark for hallucination detection at a passage level, rather than at the sentence level. 

\subsection{Techniques being used to Mitigate Hallucinations}

In \cite{b13}, a framework utilizing Reinforcement Learning from Knowledge (RLKF) is proposed to encourage LLMs to leverage their internal knowledge more effectively. Notably, the efficacy of knowledge graphs in mitigating hallucination is demonstrated in \cite{b14}. In \cite{b15}, self-reflection is employed to alleviate hallucination in a medical question-answering system. Focusing on linguistic techniques, \cite{b16} underscores the impact of correctness and formality in prompts to reduce hallucination. A novel framework, HaloCheck, is introduced in \cite{b17}, indicating the severity of hallucination in open-source LLMs with reduced parameter size. Additionally, knowledge infusion is explored to mitigate hallucination in challenging domains. \cite{b18} presents a framework employing Chain of Natural Language Inference (CoNLI) for hallucination detection and subsequent reduction through post-editing. The comprehensive survey in \cite{b19} details various techniques for mitigating hallucination, encompassing 32 different types. The paper also emphasizes the influence of the training corpus on the hallucinating behavior of these models. 

\subsection{Importance of evaluating Consistency and Reasoning in LLMs}

In \cite{b20}, the authors delineate that hallucination behavior in LLMs is influenced significantly by reasoning capabilities (i.e., the ability to learn and recall) and consistency. Notably, recent research, as evidenced by papers such as \cite{b21}, \cite{b22}, and \cite{b23}, has focused on measuring the consistency of responses generated by LLMs. This highlights the critical role of consistency in the hallucination phenomenon in LLMs, emphasizing the importance of regular evaluation, especially as these models continuously improve. The literature, including \cite{b24}, consistently indicates that LLMs often struggle with reasoning. In \cite{b25}, the authors propose a method to compel LLMs to enhance their reasoning capabilities and provide accurate explanations, particularly on sensitive topics. These endeavors signify ongoing efforts to augment the reasoning abilities of LLMs, underscoring the necessity for periodic evaluations to monitor their progress.

\section{Methodology}

For performing the experiments, the Boolq dataset is used which has 9427 instances of true and false type of questions. Each question is associated with an answer and an explanation backing it. These answers and explanations are considered as the ground truth for the experiments. All the questions in the dataset are based on general knowledge, which makes it suitable for our experiments. As all the LLMs are trained on general corpus data which includes information related to general knowledge.

\begin{table}[tb]
    \caption{Full name of the models along with their abbreviations used in the paper}
    \centering
    \begin{tabular}{|c|c|}
        \hline
        \textbf{Abbreviation} & \textbf{Full Name} \\ \hline
        G4P & gpt-4-turbo-preview \\ \hline
        G4 & gpt-4 \\ \hline
        G3.5 & gpt-3.5-turbo\\ \hline
        L2 & llama-2-7b-chat  \\ \hline
        M7 & mistral-7b-instruct\\ \hline
        Mx & mixtral-8x7b-instruct \\ 
         \hline
    \end{tabular}
    \label{tab:1}
\end{table}

The following zero-shot prompt is used for the experiment:

\begin{tcolorbox}[colback=yellow!10!white,colframe=green!75!black,title= Example Query]
  \textbf{Prompt:} Answer in True or False and give explanation for it. Question: 'Do Iran and Afghanistan speak the same language ?'\\
  \textbf{Instruction:} The format of the response should be Answer: ..... Explanation: ....... If you don't know the answer just say 'skip'.   
  \tcblower
  \textbf{Response:} Answer: False \\
Explanation: While there is some overlapping in the languages that are spoken in Iran and Afghanistan, they are not the same. The official language of Iran is Persian (also known as Farsi), while Afghanistan has two official languages - Pashto and Dari. Dari is a variety of the Persian language and is often mutually intelligible with Farsi but it's not the same. Moreover, numerous regional languages and dialects exist in both countries.  
\end{tcolorbox}

In the above example, the question is taken from the dataset and included in the prompt. The generated response from the LLM is then compared with the answer and explanation provided for each question in the dataset to assess the LLM's performance.

\subsection{Evaluating Consistency of the LLMs}

The consistency of the LLMs is evaluated by giving the same query to the models three times. If the response of the LLMs (considering only true and false) don't change, then we say that the LLM is consistent over that particular query else it is inconsistent.

The consistency of the LLMs is measured by measuring the consistency and inconsistency percentage. The Consistency percentage is given by:
\begin{equation}
    C_p = \frac{n_c}{n_t} \times 100 \label{eq1}
\end{equation}

and the inconsistency percentage is given by:
\begin{equation}
    C_n = \frac{n_{ic}}{n_t} \times 100 \label{eq2}
\end{equation}
where, 

\(C_p\) is the consistency percentage, 

\(C_n\) is the inconsistency percentage, 

\(n_c\) is the total number of consistent responses, 

\(n_{ic}\) is the total number of inconsistent responses, and 

\(n_t\) is the total number of questions in the dataset.

Along with this the skip percentage of the models is also measured. The skip percentage is the total number of times a Public or a Proprietary LLM skips the query upon the total number of queries. It is represented as:
\begin{equation}
    S_p = \frac{n_q}{n_t} \times 100 \label{eq3}
\end{equation}
where,

\(S_p\) is the skip percentage, and 

\(n_q\) is the total number of queries where the model gave the response as 'skip'.
\subsection{Evaluating Reasoning Capabilities of the LLMs}

In order to evaluate the reasoning capabilities of the LLMs, the generated explanation of the LLM is compared with the ground truth (the explanation in the dataset).

The following metrics are used in order to compare the similarity between the generated response and the ground truth:
\begin{enumerate}
    \item \textbf{BERT Score}: The BERT Score takes the semantics of the generated response (by the Public and Proprietary LLMs) and the ground truth into consideration. Based on this it tells us the similarity between the two texts.
    \item \textbf{BLEU Score}: The BLUE Score instead of taking the semantics of the generated response and the ground truth, gives us a score on clipped n-gram match. Considering both BERT and BLEU score gives us a wider picture of the relationship between the two texts.
    \item \textbf{F-1 Score}: The F-1 score gives us a balanced measure between the amount of information present in the generated response that is actually found in the ground truth and the amount of key information present in the generated response when compared with the ground truth. 
\end{enumerate}

These metrics helps to get an idea of whether these LLMs are able to generate correct explanations for their generate answers. And consequently, whether these LLMs are able to reason or not.  

\section{Result and Analysis}

Table \ref{tab:1} gives the full names of the models with their respective abbreviations that are used throughout the research paper. Table \ref{tab:2} gives the consistency, inconsistency and skip percentage of different public and proprietary LLMs. It shows a complete picture of how different models perform in terms of consistency when provided with same queries. Fig. \ref{fig:Consistency} shows the visualization of the consistency and skip percentage of different models.  

 \usepgfplotslibrary{colormaps, groupplots}%
 \pgfplotsset{compat=1.10}
 \pgfplotsset{
    colormap={slategraywhite}{
        rgb255=(112,128,144)
        rgb255=(255,159,101)
    }}

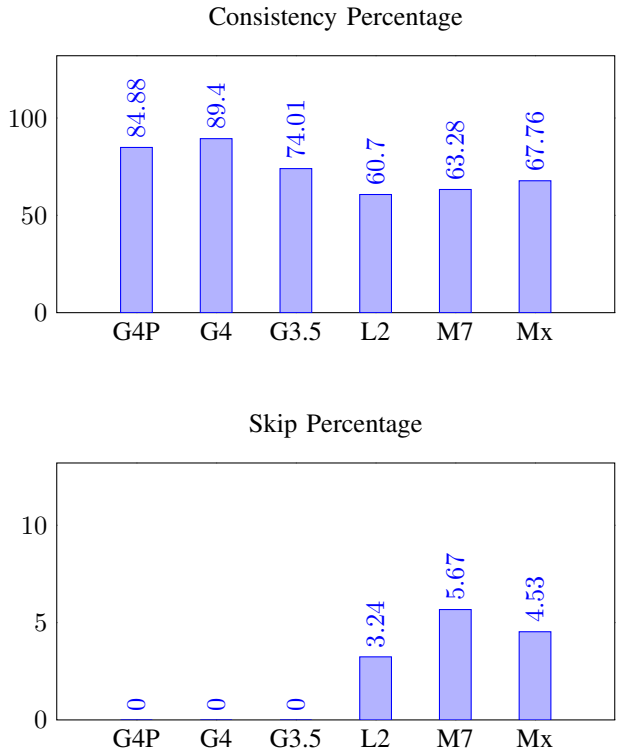
\begin{figure}[tb]
\begin{tikzpicture}
    \begin{groupplot}[group style={group size=1 by 2, vertical sep=2cm,},width=9cm,height=5cm]
        \nextgroupplot[
            symbolic x coords={G4P, G4, G3.5, L2, M7, Mx},
            major tick length=0cm,
            xtick=data,
            ymin=0.0, ymax=110,
            enlarge x limits=0.2,
            enlarge y limits={upper,value=0.2},
            nodes near coords,
            ybar,
            every node near coord/.append style={rotate=90, anchor=west},
            bar width = 12pt,
            title = {Consistency Percentage}
        ]

        \addplot coordinates {(G4P,84.88) (G4,89.40) (G3.5,74.01) (L2,60.70) (M7, 63.28) (Mx,67.76)};

        \nextgroupplot[
            symbolic x coords={G4P, G4, G3.5, L2, M7, Mx},
            major tick length=0cm,
            xtick=data,
            ymin=0.0, ymax=11,
            enlarge x limits=0.2,
            enlarge y limits={upper,value=0.2},
            nodes near coords,
            ybar,
            every node near coord/.append style={rotate=90, anchor=west},
            bar width = 12pt,
            title = {Skip Percentage}
        ]

        \addplot coordinates {(G4P,0.00) (G4,0.00) (G3.5,0.00) (L2,3.24) (M7,5.67) (Mx,4.53)};
\end{groupplot}
\end{tikzpicture}
\caption{Consistency and Skip Percentage of public and proprietary models}
\label{fig:Consistency}
\end{figure}

Table \ref{tab:3} gives the BLEU and F-1 scores of different public and proprietary LLMs. And, Fig. \ref{fig:Reasoning} gives us the visualization of these scores for different models. The BERT Score is not included in the result as it remained almost similar for all the models (about 0.85).

\begin{table}[tb]
    \caption{Consistency Percentage and Inconsistency Percentage for different LLMs}
    \begin{center}
    \begin{tabular}{|c|c|c|c|}
    \hline
        \textbf{Models} & \textbf{Consistency \%} &\textbf{Inconsistency \%} &\textbf{Skip \%} \\ \hline
         G4P&84.88 &15.11 &0.00 \\
         G4&89.40 &10.60 &0.00 \\
         G3.5&74.01 &25.99 &0.00 \\ \hline
         L2&60.70 &39.30 &3.24 \\
         M7&63.28 &36.72 &5.67 \\
         Mx&67.76 &32.24 &4.53 \\
         \hline
    \end{tabular}
    \label{tab:2}
    \end{center}
\end{table}

\begin{table}[tb]
    \caption{BLEU and F-1 Scores for different public and proprietary LLMs}
    \centering
    \begin{tabular}{|c|c|c|}
    \hline
        \textbf{Model} &\textbf{BLEU Score} &\textbf{F-1 Score}  \\ \hline
        G4P&0.016 &0.284 \\
         G4&0.033 &0.320 \\
         G3.5&0.037 &0.318 \\ \hline
         L2&0.024 &0.296 \\
         M7&0.031 &0.301 \\
         Mx&0.048 &0.322 \\
         \hline
    \end{tabular}
    \label{tab:3}
\end{table}

 \usepgfplotslibrary{colormaps, groupplots}%
 \pgfplotsset{compat=1.10}
 \pgfplotsset{
    colormap={slategraywhite}{
        rgb255=(112,128,144)
        rgb255=(255,159,101)
    }}

\begin{figure}[!htbp]
\begin{tikzpicture}
    \begin{groupplot}[group style={group size=1 by 2, vertical sep=2cm,},width=9cm,height=5cm]
        \nextgroupplot[
            symbolic x coords={G4P, G4, G3.5, L2, M7, Mx},
            major tick length=0cm,
            xtick=data,
            ymin=0.0, ymax=0.1,
            enlarge x limits=0.2,
            enlarge y limits={upper,value=0.2},
            nodes near coords,
            ybar,
            every node near coord/.append style={rotate=90, anchor=west},
            bar width = 12pt,
            title = {BLEU Score}
        ]

        \addplot coordinates {(G4P,0.016) (G4,0.033) (G3.5,0.037) (L2,0.024) (M7, 0.031) (Mx, 0.048)};

        \nextgroupplot[
            symbolic x coords={G4P, G4, G3.5, L2, M7, Mx},
            major tick length=0cm,
            xtick=data,
            ymin=0.0, ymax=0.5,
            enlarge x limits=0.2,
            enlarge y limits={upper,value=0.2},
            nodes near coords,
            ybar,
            every node near coord/.append style={rotate=90, anchor=west},
            bar width = 12pt,
            title = {F-1 Score}
        ]

        \addplot coordinates {(G4P, 0.284) (G4, 0.320) (G3.5, 0.318) (L2,0.296) (M7,0.301) (Mx,0.322)};
\end{groupplot}
\end{tikzpicture}
\caption{BLEU and F-1 scores for public and proprietary models}
\label{fig:Reasoning}
\end{figure}
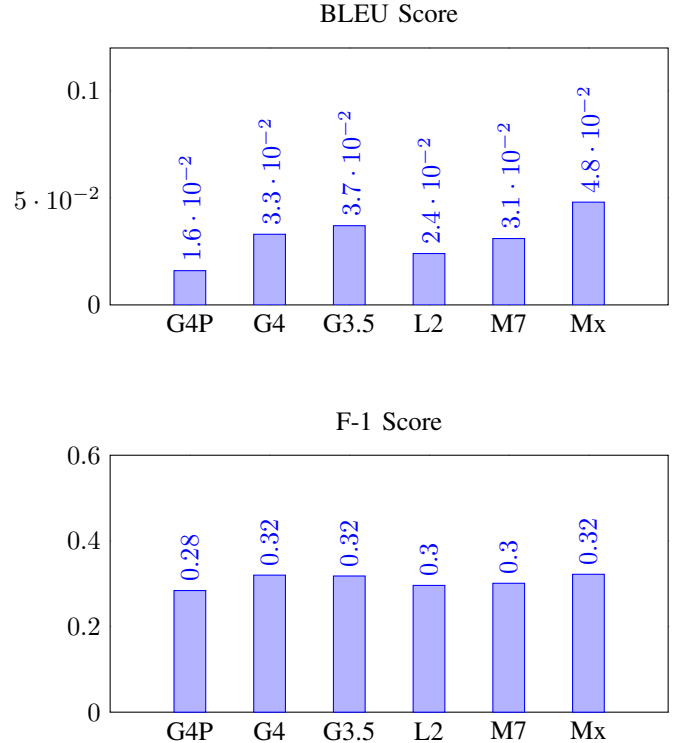

\subsection{Findings}

It's important to highlight that the inconsistency observed in both public and proprietary models occurred when they provided incorrect answers initially. Furthermore, they consistently provided incorrect explanations (reasoning) alongside these wrong answers. This might mean that these models have less context about the questions when they provide wrong answers. Plus, there is a direct relationship that exists between the consistency and reasoning capabilities of these models. Moreover, these models don't 'skip' the queries/questions and tend to give wrong answers even after explicitly mentioning in the prompt. This shows that there is some inherent reasoning issues in these models.

Some of the other findings that can be inferred from the results are as follows:
\begin{itemize}
    \item All proprietary models perform better than public models in terms of consistency. This shows that the hallucination is less in proprietary models than in public models.
    \item Among the proprietary models, GPT-4-Turbo-Preview performs the worst in both consistency and reasoning capabilities.
    \item Overall LLAMA-2-7b model performs worst in both consistency and reasoning capabilities.
    \item The BLEU score of all the models is very less while the BERT score of all the models is quite high. This shows that even though the generated explanation and the ground truth explanation are semantically similar but their n-gram similarity is very less.
    \item Because of the less values of the BLEU score and almost similar values of the BERT score, the F-1 score of the models doesn't show much variation.  
\end{itemize}

\section{Conclusion}
It is evident from the results that the language models show inconsistencies only when giving wrong answers which are accompanied by incorrect explanations. This shows that these models posses a lack of contextual awareness when providing incorrect responses.

Based on this study, it is clear that the current language models have inherent reasoning issues. These issues need to be fixed in order to achieve better performance and reliability in natural language processing tasks. 


\begin{thebibliography}{00}
\bibitem{huang2022towards} Huang, Jie, and Kevin Chen-Chuan Chang. "Towards reasoning in large language models: A survey." arXiv preprint arXiv:2212.10403 (2022).
\bibitem{b2} Clark, Christopher, et al. "BoolQ: Exploring the surprising difficulty of natural yes/no questions." arXiv preprint arXiv:1905.10044 (2019).
\bibitem{b3} Touvron, Hugo, et al. "Llama 2: Open foundation and fine-tuned chat models." arXiv preprint arXiv:2307.09288 (2023).
\bibitem{b4} Jiang, Albert Q., et al. "Mistral 7B." arXiv preprint arXiv:2310.06825 (2023).
\bibitem{b5} Jiang, Albert Q., et al. "Mixtral of experts." arXiv preprint arXiv:2401.04088 (2024).
\bibitem{b6} Rawte, Vipula, Amit Sheth, and Amitava Das. "A survey of hallucination in large foundation models." arXiv preprint arXiv:2309.05922 (2023).
\bibitem{b7} Curran, Shawn, Sam Lansley, and Oliver Bethell. "Hallucination is the last thing you need." arXiv preprint arXiv:2306.11520 (2023).
\bibitem{b8} Xu, Ziwei, Sanjay Jain, and Mohan Kankanhalli. "Hallucination is inevitable: An innate limitation of large language models." arXiv preprint arXiv:2401.11817 (2024).
\bibitem{b9} Stringhi, Elisabetta. "Hallucinating (or poorly fed) LLMs? The problem of data accuracy." i-lex 16.2 (2023): 54-63.
\bibitem{b10} Rawte, Vipula, et al. "The Troubling Emergence of Hallucination in Large Language Models--An Extensive Definition, Quantification, and Prescriptive Remediations." arXiv preprint arXiv:2310.04988 (2023).
\bibitem{b11} Chen, Yuyan, et al. "Hallucination detection: Robustly discerning reliable answers in large language models." Proceedings of the 32nd ACM International Conference on Information and Knowledge Management. 2023.
\bibitem{b12} Sadat, Mobashir, et al. "Delucionqa: Detecting hallucinations in domain-specific question answering." arXiv preprint arXiv:2312.05200 (2023).
\bibitem{b13} Yang, Shiping, Renliang Sun, and Xiaojun Wan. "A new benchmark and reverse validation method for passage-level hallucination detection." arXiv preprint arXiv:2310.06498 (2023).
\bibitem{b14} Liang, Yuxin, et al. "Learning to trust your feelings: Leveraging self-awareness in llms for hallucination mitigation." arXiv preprint arXiv:2401.15449 (2024).
\bibitem{b15} Agrawal, Garima, et al. "Can knowledge graphs reduce hallucinations in LLMs?: A survey." arXiv preprint arXiv:2311.07914 (2023).
\bibitem{b16} Ji, Ziwei, et al. "Towards mitigating LLM hallucination via self reflection." The 2023 Conference on Empirical Methods in Natural Language Processing. 2023.
\bibitem{b17} Rawte, Vipula, et al. "Exploring the relationship between llm hallucinations and prompt linguistic nuances: Readability, formality, and concreteness." arXiv preprint arXiv:2309.11064 (2023).
\bibitem{b18} Elaraby, Mohamed, et al. "Halo: Estimation and reduction of hallucinations in open-source weak large language models." arXiv preprint arXiv:2308.11764 (2023).
\bibitem{b19} Lei, Deren, et al. "Chain of Natural Language Inference for Reducing Large Language Model Ungrounded Hallucinations." arXiv preprint arXiv:2310.03951 (2023).
\bibitem{b20} Huang, Lei, et al. "A survey on hallucination in large language models: Principles, taxonomy, challenges, and open questions." arXiv preprint arXiv:2311.05232 (2023).
\bibitem{b21} Chen, Xinyun, et al. "Universal self-consistency for large language model generation." arXiv preprint arXiv:2311.17311 (2023).
\bibitem{b22} Ye, Wentao, et al. "Assessing Hidden Risks of LLMs: An Empirical Study on Robustness, Consistency, and Credibility." arXiv preprint arXiv:2305.10235 (2023).
\bibitem{b23} Alexander, Rohan, et al. "Evaluating the Decency and Consistency of Data Validation Tests Generated by LLMs." arXiv preprint arXiv:2310.01402 (2023).
\bibitem{b24} Valmeekam, Karthik, et al. "Large Language Models Still Can't Plan (A Benchmark for LLMs on Planning and Reasoning about Change)." arXiv preprint arXiv:2206.10498 (2022).
\bibitem{b25} Kenny, Eoin, and Julie Shah. "In Pursuit of Regulatable LLMs." NeurIPS 2023 Workshop on Regulatable ML. 2023.
\end{thebibliography}
\end{document}